\documentclass{article}

\usepackage[accepted]{icml2026}
\setcitestyle{numbers,square}

\usepackage[T1]{fontenc}
\usepackage[utf8]{inputenc}
\usepackage{microtype}
\usepackage{graphicx}
\usepackage{booktabs}
\usepackage{amsmath,amssymb}
\usepackage{subcaption}
\usepackage{algorithm}
\usepackage{algpseudocode}
\usepackage{hyperref}
\usepackage{bbm}

\icmltitlerunning{Task-Aligned Stability Analysis for VLM Hazard Detection}

\begin{document}

\twocolumn[
\icmltitle{Task-Aligned Stability Analysis of Vision-Language Models for Autonomous Driving Hazard Detection}

\begin{icmlauthorlist}
\icmlauthor{Everett Richards}{sdsu}
\end{icmlauthorlist}

\icmlaffiliation{sdsu}{Department of Computer Science, San Diego State University}

\icmlcorrespondingauthor{Everett Richards}{ehrichards9@gmail.com}
\vskip 0.3in
]

\printAffiliationsAndNotice{}

\begin{abstract}
Vision-language models (VLMs) are increasingly used for scene understanding in autonomous driving, but robustness analysis often relies on task-agnostic embedding stability alone.
We study whether corruption-induced embedding drift predicts changes in a task-aligned hazard score derived from CLIP image-text similarities.
Using controlled corruptions on BDD100K road scenes, we compare embedding drift against margin drift, defined as the change in hazard score under perturbation.
The relationship is highly corruption-dependent: some families exhibit strong coupling between representation drift and decision drift, while others induce hazardous decision instability despite relatively modest embedding change.
Furthermore, corruption families differ in failure direction: most suppress hazard detections via false negatives, while occlusion instead triggers false alarms, suggesting that benchmark design should account for asymmetric failure modes, not just overall instability rates.
These results suggest that robustness benchmarks should include task-aligned stability measures in addition to embedding-level perturbation statistics.
\end{abstract}

\vspace{-0.5cm}

\section{Introduction}

Autonomous Vehicles (AVs) are becoming increasingly popular, and are already being deployed in many cities around the world~\cite{litman2025avpredictions}. However, public confidence in self-driving technology remains astonishingly low, with a 2025 survey suggesting that only 38\% of Americans would be willing to ride in a self-driving car if given the opportunity \cite{WALI2026104941}. Furthermore, a 2024 study demonstrated that people are more inclined to blame a self-driving car than a human-operated car for a collision, even in otherwise equivalent scenarios \cite{ZHANG2024103887}.

In recent years, researchers have identified perception as an essential safety factor in autonomous driving~\cite{richards2025_ecod}.
Autonomous vehicles depend on perception systems that remain reliable under distribution shift.
 Hazard detection is a particularly safety-critical instance of this problem because small visual changes can alter the driving decision implied by a scene.
 Vision-language models (VLMs) are attractive for this setting because they combine visual recognition with prompt-based semantic reasoning, enabling natural-language descriptions of hazardous and non-hazardous traffic scenarios~\cite{bordes2024introductionvisionlanguagemodeling,Xia_2026_CVPR}.
 However, prior work shows that VLM-based hazard detectors can be brittle, overreact in ambiguous settings, and impose substantial computational cost~\cite{choi2025bettersafesorryoverreaction,adil2025usingvisionlanguagemodels}.

A common robustness perspective is to measure how much the image embedding changes under corruption. While useful, this quantity is task-agnostic: a large change in embedding space need not produce a large change in the decision variable that matters for safety. We therefore contrast \emph{embedding drift} with a task-aligned \emph{margin drift} based on a CLIP hazard score. This aligns with broader efforts toward quantitative reliability analysis, robustness under distribution shift, and principled benchmark design.

Whereas prior works have explored the effects of data perturbations on multi-modal object detection pipelines~\cite{richards2025_chaostoclarity} and reinforcement learning-based action trajectory planning~\cite{richards2025modeling}, this paper aims to model the degradation of VLM-based hazard detection and contribute to our understanding of notable failure cases.

Our central question is simple: \emph{to what extent does corruption-induced embedding drift predict decision-relevant instability in VLM hazard detection?} We find that the answer depends strongly on corruption family. JPEG compression and downsampling produce relatively predictable behavior, whereas motion blur and occlusion can trigger high decision flip rates without proportionally large embedding movement. This suggests that representation stability alone is an incomplete proxy for safety-relevant robustness.

\section{Method}
\paragraph{Hazard scoring.}
Let $\hat f(x)$ and $\hat g(t)$ denote frozen CLIP image and text encoders (both normalized). We define hazardous prompts $\mathcal{H}$ and non-hazardous prompts $\mathcal{N}$, then score image $x$ by
\begin{equation}
\label{eq:define_lambda}
\Lambda(x)=\max_{h\in \mathcal{H}}\langle \hat f(x),\hat g(h)\rangle-\max_{n\in \mathcal{N}}\langle \hat f(x),\hat g(n)\rangle,
\end{equation}
where larger values indicate greater hazard likelihood. Positive scores correspond to hazardous scenes and negative scores to non-hazardous scenes.

Our prompt sets are intentionally simple and interpretable. Hazardous prompts include phrases such as \emph{a pedestrian crossing the road}, \emph{a cyclist in the lane}, and \emph{a stopped vehicle blocking traffic}. Non-hazardous prompts include \emph{a clear road with no obstacles}, \emph{a normal highway scene}, and \emph{a clear lane ahead}. This follows the prompt-based hazard scoring strategy used in recent autonomous-driving work built on CLIP~\cite{CLIPradford2021learningtransferablevisualmodels,greer2026visionlanguagenovelrepresentations}.

\paragraph{Corruptions and drift metrics.}
For corruption family~$c$ and severity $\alpha\in\{1,\dots,5\}$, let $T_{c,\alpha}(x)$ denote the corrupted image. We evaluate seven families: fog, Gaussian blur, motion blur, JPEG compression, low light, occlusion, and downsampling. We then define
\begin{equation}
\Delta(x,c,\alpha)=\|\hat f(x)-\hat f(T_{c,\alpha}(x))\|_2
\end{equation}
and
\begin{equation}
D_{\Lambda}(x,c,\alpha)=|\Lambda(x)-\Lambda(T_{c,\alpha}(x))|.
\end{equation}
Here $\Delta$ is task-agnostic embedding drift and $D_{\Lambda}$ is task-aligned margin drift.

We also record a binary \emph{flip} indicator,
\begin{equation}
\texttt{flip}=\mathbbm{1}[\mathrm{sign}(\Lambda(x))\neq \mathrm{sign}(\Lambda(T_{c,\alpha}(x)))],
\end{equation}
which captures whether corruption changes the hazard verdict.

\begin{algorithm}[t]
\caption{Task-aligned corruption evaluation}
\label{alg:evaluation}
\begin{algorithmic}[1]
\Require Images $\mathcal X$, corruptions $\mathcal C$, severities $\mathcal A$, prompt collections $\mathcal{H},\mathcal{N}$
\For{$c\in\mathcal C$}
  \For{$\alpha\in\mathcal A$}
    \For{$x\in\mathcal X$}
      \State Compute $\Lambda(x)$ and $\hat f(x)$
      \State Form $x' = T_{c,\alpha}(x)$
      \State Compute $\Lambda(x')$ and $\hat f(x')$
      \State Set $\Delta=\|\hat f(x)-\hat f(x')\|_2$
      \State Set $D_{\Lambda}=|\Lambda(x)-\Lambda(x')|$
      \State Set $\texttt{flip}=\mathbbm{1}[\mathrm{sign}(\Lambda(x))\neq \mathrm{sign}(\Lambda(x'))]$
    \EndFor
  \EndFor
\EndFor
\end{algorithmic}
\end{algorithm}

\paragraph{Cauchy-Schwarz upper bound.} Suppose that the most-aligned prompts $h^* \in \mathcal{H}$ and $n^* \in \mathcal{N}$ are stable under perturbation, i.e. that their relative order is unchanged by corruption operators. Then the $\Lambda$ formulation in Eq.~\ref{eq:define_lambda} collapses to:
\begin{equation}
    \Lambda(x) = \langle \hat{f}(x), \hat{g}(h^*) - \hat{g}(n^*) \rangle
\end{equation}
Then, the Cauchy-Schwarz inequality gives us an upper bound on margin drift $D_\Lambda$, where the coefficient $L$ depends on the prompt sets $\mathcal{H},\mathcal{N}$:
\begin{equation}
    \label{eq:upper_bound}
    |D_\Lambda| \leq \| \hat{g}(h^*) - \hat{g}(n^*) \|_2 \cdot \| \hat{f}(x) - \hat{f}(T(x)) \|_2 = L \cdot \Delta
\end{equation}

\section{Experimental setup}
We use 2,000 images sampled from the BDD100K validation split~\cite{bdd100kyu2020bdd100kdiversedrivingdataset}. For each image, corruption family, and severity, we compute the metrics above and then summarize each family using: (i) Spearman correlation $\rho(\Delta,D_{\Lambda})$, (ii) flip rate, and (iii) mean drift as a function of severity. This design isolates the relationship between representation change and decision instability under controlled shift.

\section{Results}
\subsection{Hazard Score Behavior}
The CLIP-based score produces plausible rankings: low-scoring scenes are typically open roads with minimal interaction, while high-scoring scenes contain dense traffic, nearby actors, or explicit obstructions. This supports the use of $\Lambda$ as a continuous decision variable for analyzing corruption effects, even in the absence of a fully supervised detection benchmark. See Appendix~\ref{sec:lambda_appx} for a visual illustration of $\Lambda$ correctly approximating hazard presence in real-world driving scenes.

\subsection{Embedding Drift and Margin Drift Are Correlated, but Not Interchangeable}
Table~\ref{tab:main-results} shows strong heterogeneity across corruption families. JPEG compression and downsampling have relatively strong rank correlation between embedding drift and margin drift, suggesting that representation movement is a reasonable proxy for decision movement in those cases. In contrast, occlusion and fog show much weaker coupling. Most notably, motion blur exhibits the highest flip rate even though its drift correlation is only moderate. Thus, the corruption families that most threaten decision reliability are not necessarily those that induce the largest embedding perturbations.

\newpage



\begin{table*}[t]
\caption{
Family-wise coupling between embedding drift and decision drift, with 95\% confidence intervals on Spearman ($\rho$) and Pearson ($r$) correlation coefficients. FPR and FNR denote negative-to-positive and positive-to-negative hazard verdict flips, respectively. $\mathcal{H}$ and $\mathcal{N}$ stability measure the proportion of images whose leading prompt is unchanged under corruption.
}
\label{tab:main-results}
\centering
\setlength{\tabcolsep}{5pt}
\small
\begin{tabular}{lcccccccc}
\toprule
Corruption family & $\rho(\Delta,D_{\Lambda})$ & $r(\Delta,D_\Lambda)$ & Flip rate (\%) & FPR (\%) & FNR (\%) & $\mathcal{H}$ stability (\%) & $\mathcal{N}$ stability (\%) \\
\midrule
Downsampling & $0.615 \pm 0.035$ & $0.572 \pm 0.034$ & 5.9 & 1.8 & 4.1 & 84.2 & 74.7 \\
Fog & $0.348 \pm 0.044$ & $0.360 \pm 0.044$ & 10.8 & 5.8 & 5.1 & 94.6 & 73.4 \\
Gaussian blur & $0.587 \pm 0.038$ & $0.563 \pm 0.035$ & 16.2 & 2.7 & 13.5 & 39.4 & 55.6 \\
JPEG compression & $0.711 \pm 0.027$ & $0.618 \pm 0.031$ & 10.8 & 0.7 & 10.2 & 75.8 & 71.7 \\
Low light & $0.553 \pm 0.037$ & $0.524 \pm 0.037$ & 15.4 & 3.4 & 12.0 & 93.6 & 71.0 \\
Motion blur & $0.455 \pm 0.042$ & $0.451 \pm 0.041$ & 24.2 & 4.2 & 20.1 & 30.6 & 38.0 \\
Occlusion & $0.266 \pm 0.049$ & $0.283 \pm 0.047$ & 17.8 & 15.3 & 2.5 & 90.2 & 65.3 \\
\midrule
Average & $0.473 \pm 0.015$ & $0.438 \pm 0.016$ & 14.4 & 4.8 & 9.6 & 72.6 & 64.3 \\
\bottomrule
\end{tabular}
\end{table*}

\begin{figure}[h]
    \centering
    \includegraphics[width=1.00\linewidth]{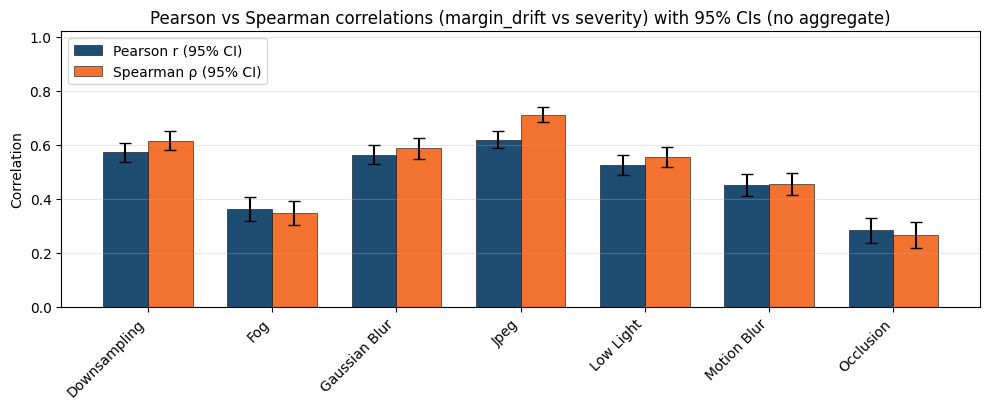}
    \caption{Family-wise comparison of Spearman correlation coefficient $\rho$ and Pearson correlation coefficient $r$, relating embedding drift $\Delta$ to margin drift $D_\Lambda$. Spearman values close to $1$ indicate a strong \textit{monotonic} relationship, whereas Pearson values close to $1$ suggest a strong \textit{linear} relationship.}
    \label{fig:spearman_pearson}
\end{figure}

\begin{figure}[h]
    \centering
    \includegraphics[width=1.00\linewidth]{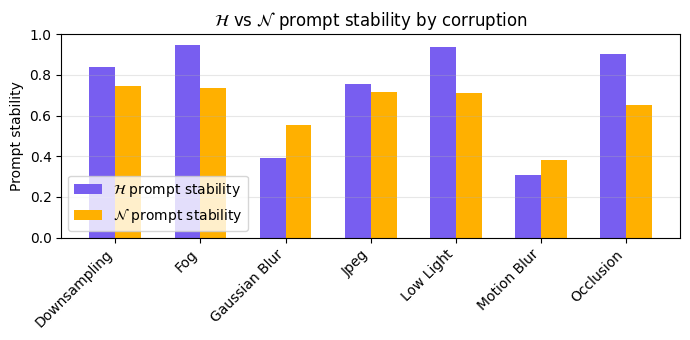}
    \caption{Family-wise comparison of $\mathcal{H}$ prompt stability and $\mathcal{N}$ prompt stability, for severity $\alpha=5$. Values close to $1$ indicate that the leading prompt is largely invariant to the corruption operator.}
    \label{fig:prompt_stability}
\end{figure}

Fig.~\ref{fig:spearman_pearson} illustrates the Spearman (rank) correlation coefficient $\rho$ and Pearson (linear) correlation coefficient $r$ for each corruption family. The two correlation metrics are tightly coupled and weakly to moderately strong across each corruption family, indicating an imperfect but non-negligible monotone and linear relationship between embedding drift and margin drift.
Fig.~\ref{fig:prompt_stability} provides a cross-family comparison of prompt stability, which measures the proportion of images whose most closely-aligned prompt in each prompt set ($\mathcal{H}$ or $\mathcal{N}$) is unchanged upon corruption. Values close to $1$ support the prompt order stability assumption required by Eq.~\ref{eq:upper_bound}, while smaller values violate this assumption. In either case, this theoretical upper bound cannot be guaranteed due to the nonlinearity in $D_\Lambda(x,c,\alpha)$ (with respect to~$\alpha$) induced by a change in leading prompt.

\begin{figure}[t]
    \centering
    \begin{subfigure}[t]{0.80\linewidth}
        \centering
        \includegraphics[width=\linewidth]{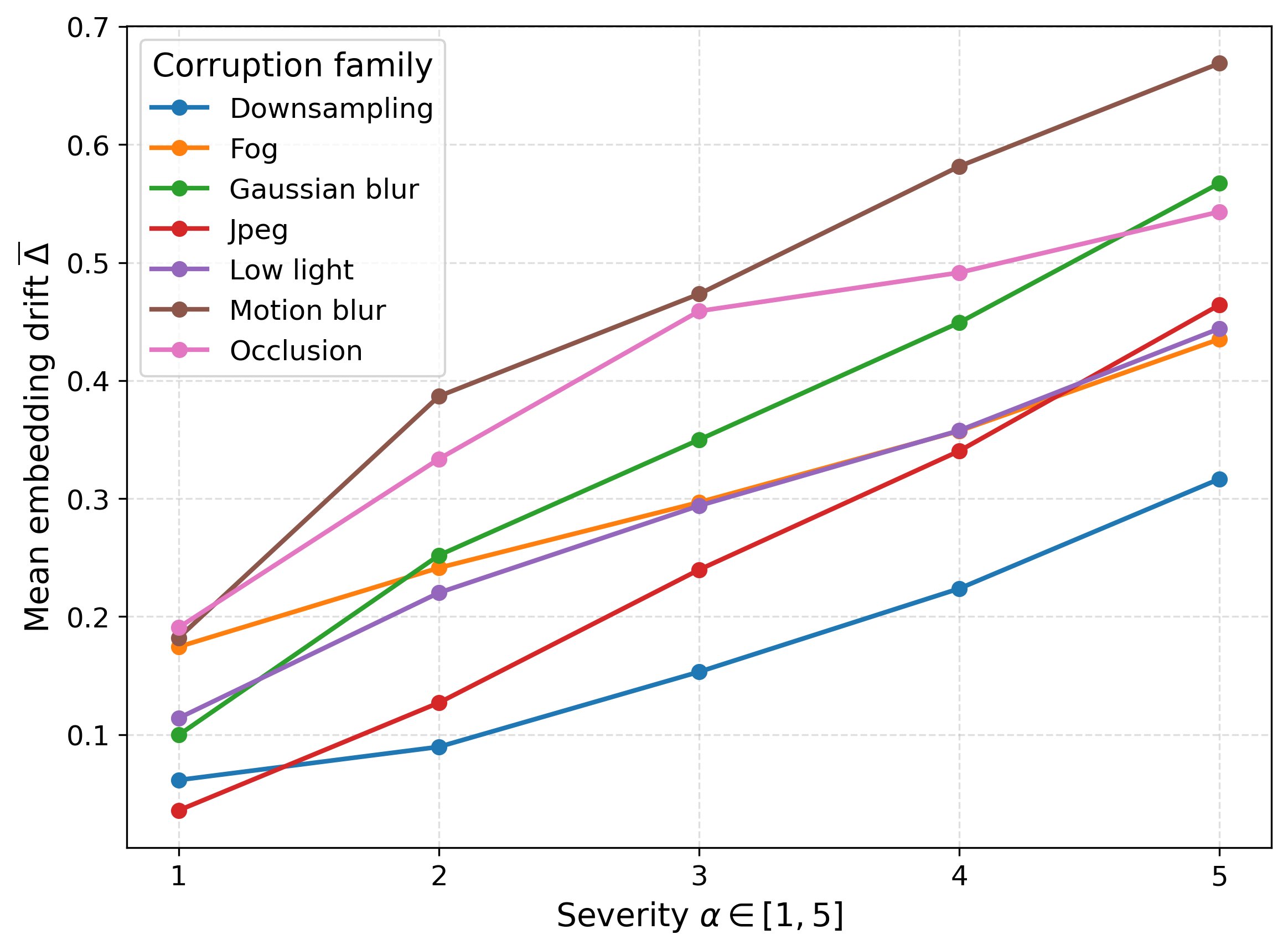}
        \caption{Mean embedding drift vs. severity.}
    \end{subfigure}
    \hfill
    \begin{subfigure}[t]{0.80\linewidth}
        \centering
        \includegraphics[width=\linewidth]{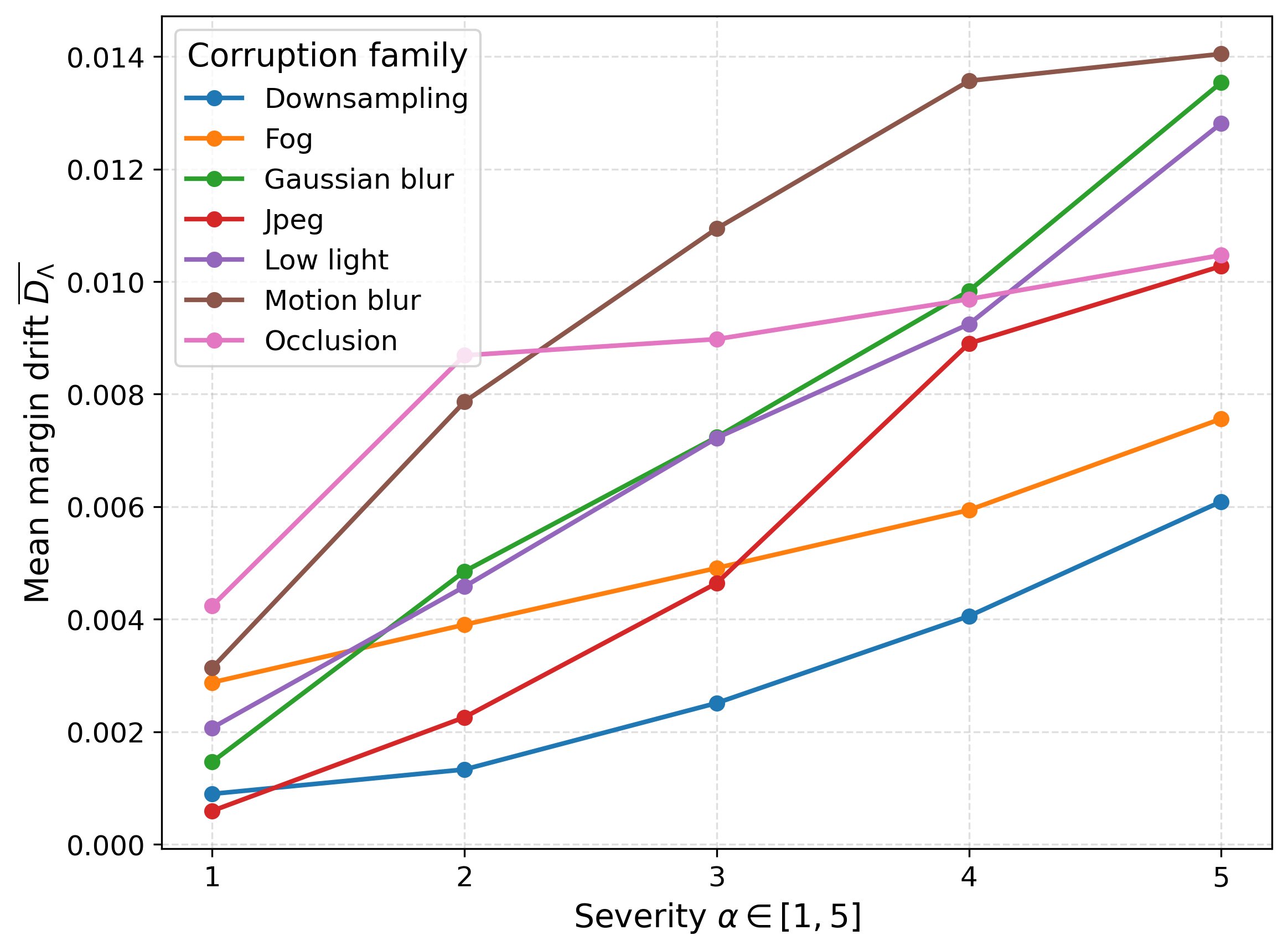}
        \caption{Mean margin drift vs. severity.}
    \end{subfigure}
    \caption{Severity trends differ across corruption families. Similar monotonicity does not imply similar safety impact.}
    \label{fig:main-fig}
\end{figure}

Figure~\ref{fig:main-fig} visually summarizes the roughly-monotone, roughly-linear relationships between $\Delta$ vs. $\alpha$ and $D_\Lambda$ vs. $\alpha$. The left panel plots mean embedding drift against severity, and the right panel plots mean margin drift against severity. Both generally increase monotonically, confirming that stronger corruptions perturb both representation and task score. However, the relative ordering differs across families. Motion blur and occlusion induce disproportionately large margin effects relative to their embedding movement, while JPEG compression causes substantial embedding change with comparatively lower decision instability.

\vspace{-0.2cm}

\subsection{Scatter Geometry of Embedding and Margin Drift}

To further analyze the relationship between representation change and decision instability, we examine the joint distribution of embedding drift $\Delta$ and margin drift $D_{\Lambda}$ across corruption families. Figure~\ref{fig:scatter_grid} visualizes this relationship. Several consistent structural patterns emerge:

\begin{figure*}[t]
    \centering

    \begin{subfigure}[t]{0.44\textwidth}
        \centering
        \includegraphics[width=\linewidth]{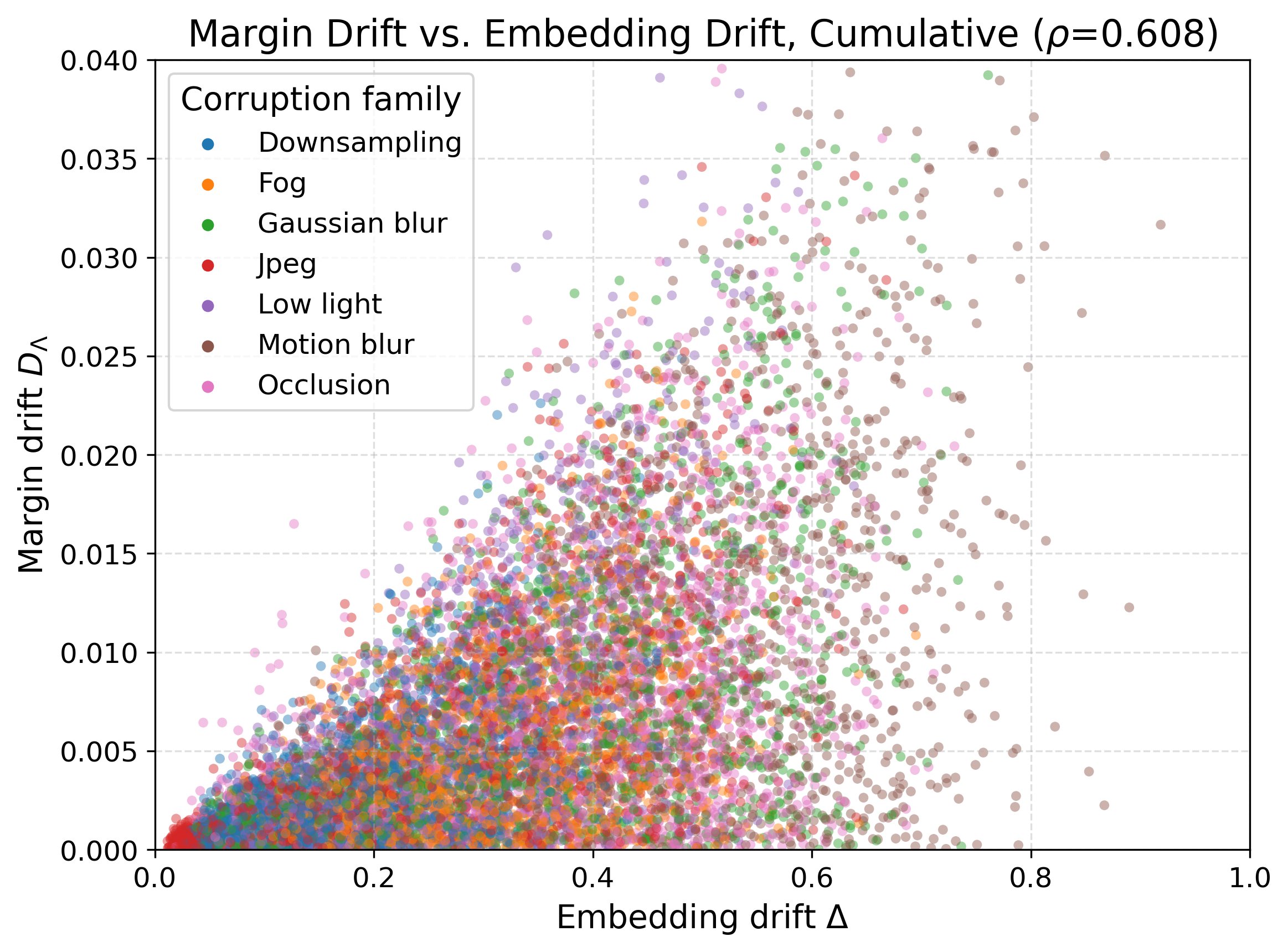}
        \caption{All corruptions}
    \end{subfigure}
    \hfill
    \begin{subfigure}[t]{0.44\textwidth}
        \centering
        \includegraphics[width=\linewidth]{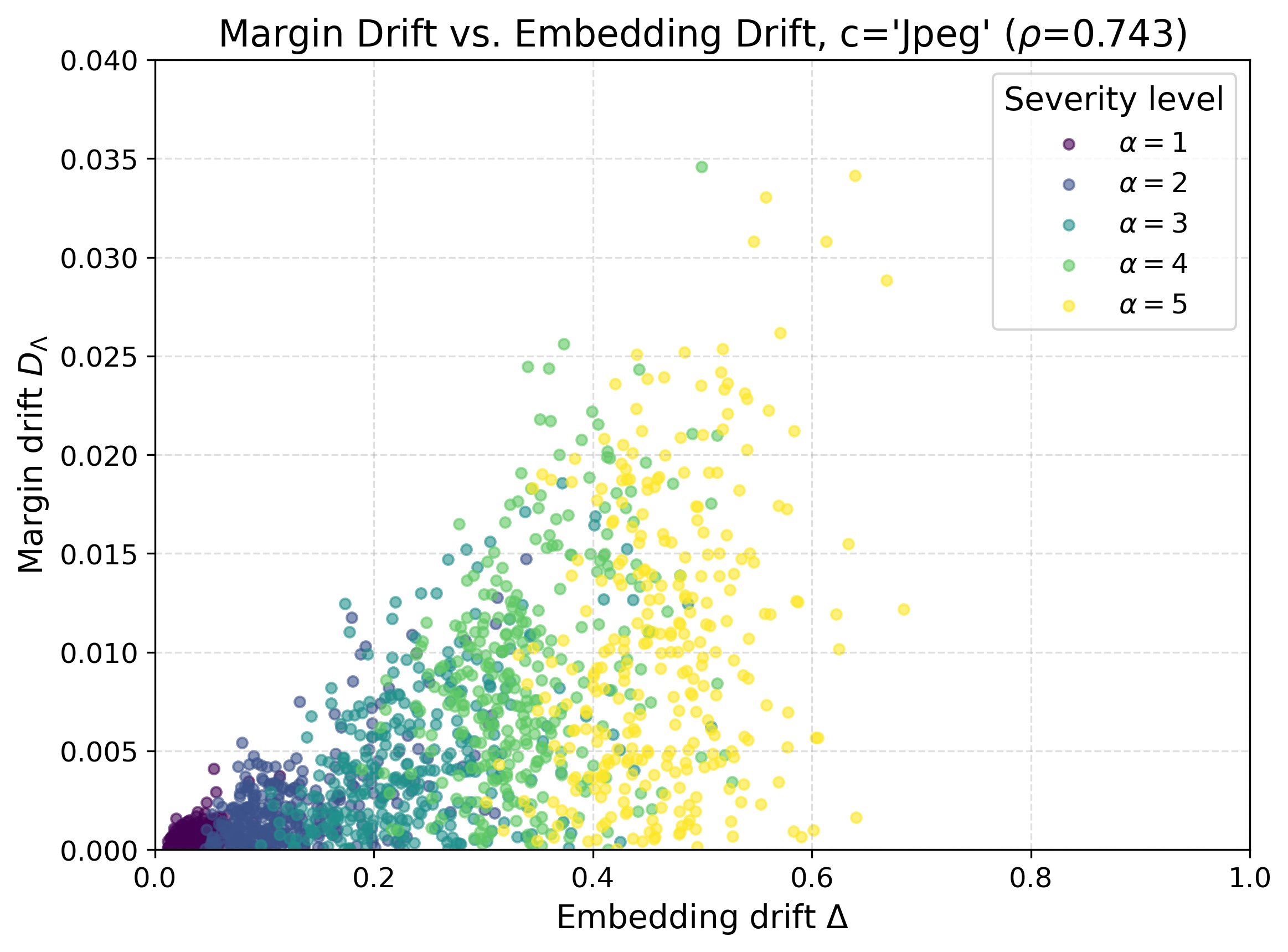}
        \caption{JPEG}
    \end{subfigure}
    \hfill
    \begin{subfigure}[t]{0.44\textwidth}
        \centering
        \includegraphics[width=\linewidth]{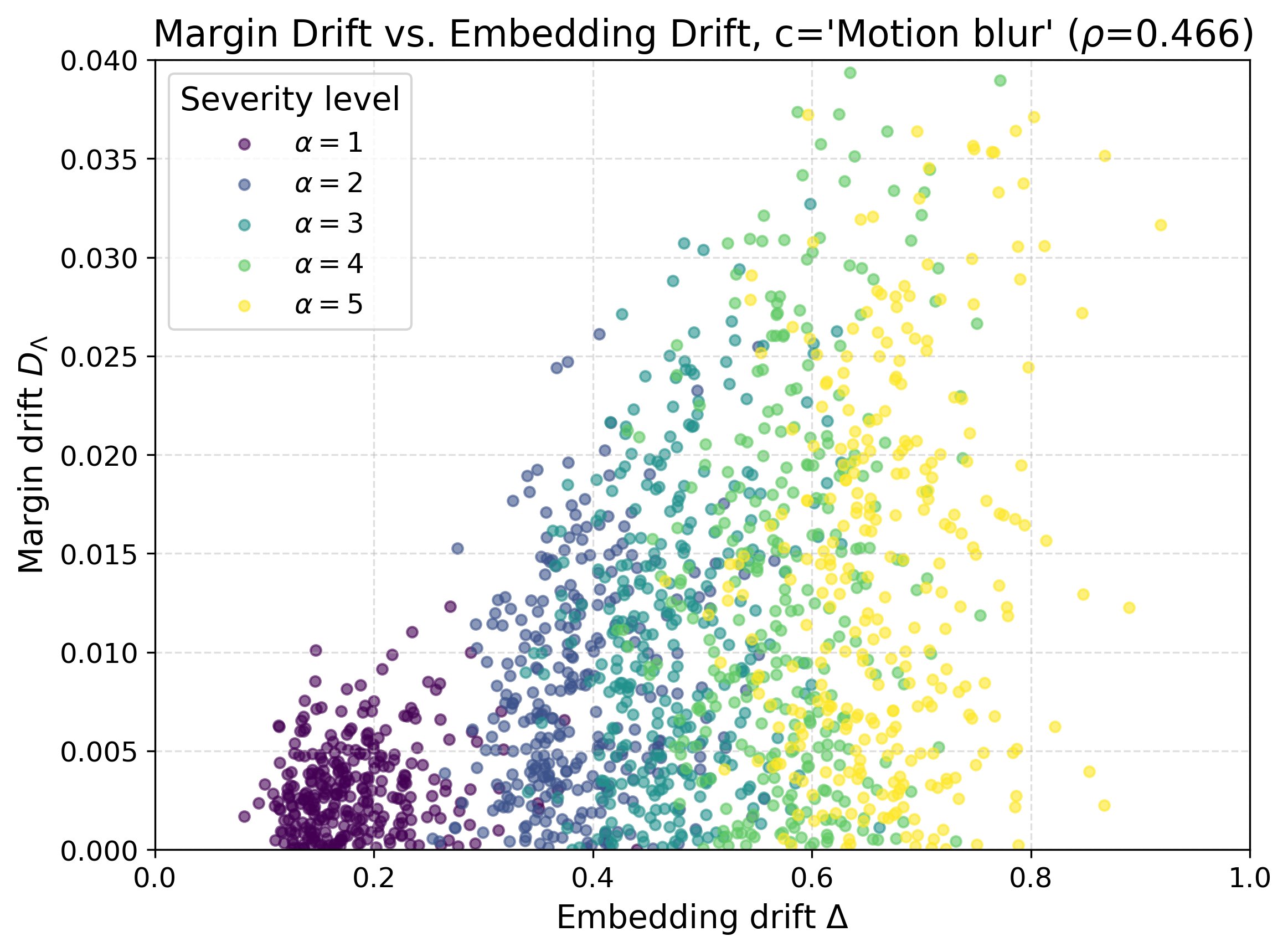}
        \caption{Motion blur}
    \end{subfigure}
    \hfill
    \begin{subfigure}[t]{0.44\textwidth}
        \centering
        \includegraphics[width=\linewidth]{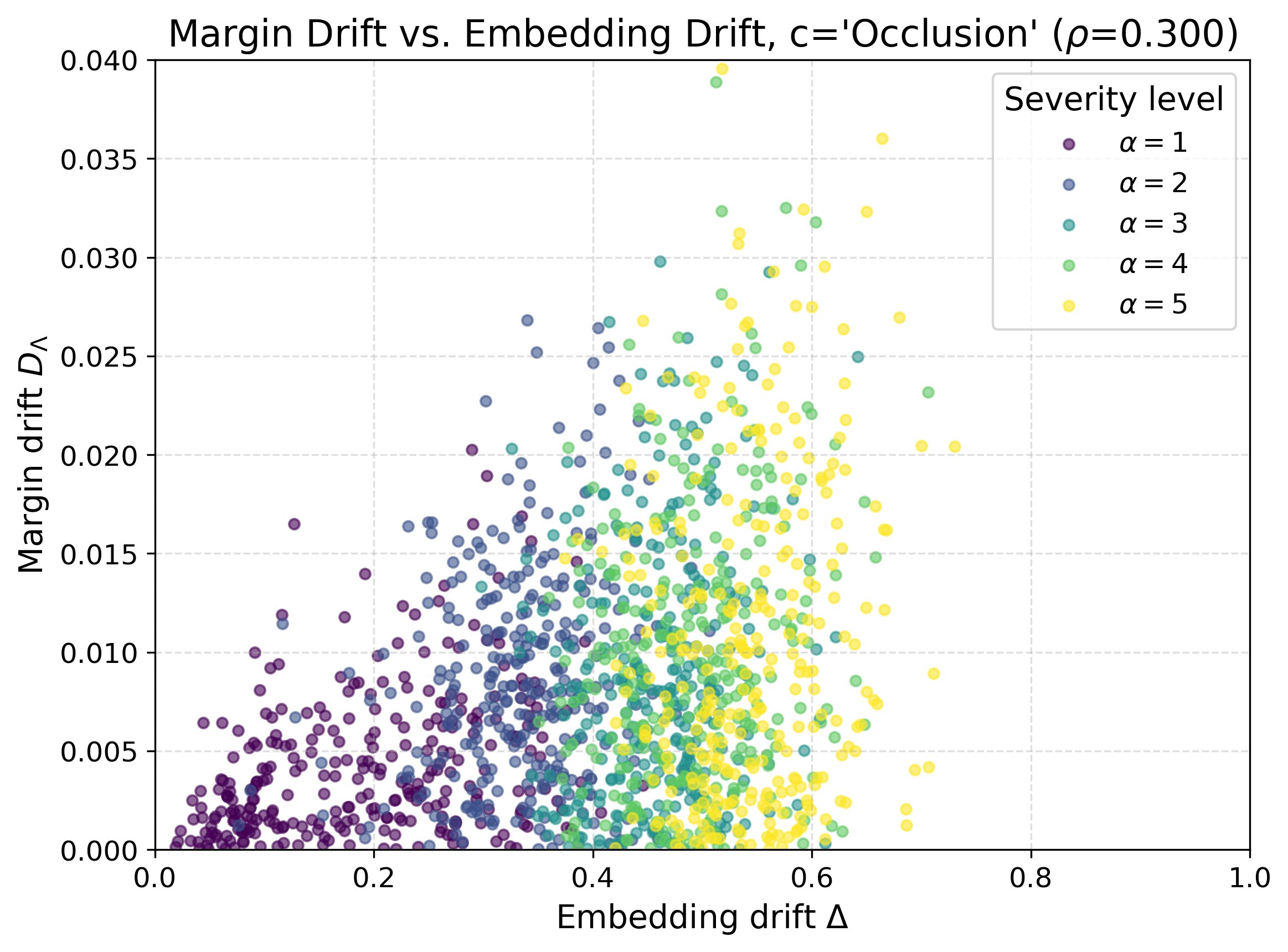}
        \caption{Occlusion}
    \end{subfigure}

    \caption{Embedding drift vs. margin drift across corruption families. Points are colored by corruption family in (a) and by severity level in (b)--(d).}
    \label{fig:scatter_grid}
\end{figure*}

\vspace{-0.2cm}

\paragraph{Upper-envelope (conical) constraint.}
Across all corruption families (Fig.~\ref{fig:scatter_grid}a), the data occupies a cone-shaped region with a clear upper boundary relating $D_{\Lambda}$ to $\Delta$. Large margin shifts are only observed when embedding drift is non-negligible, suggesting that representation change is a necessary condition for decision change. However, the spread within this cone is substantial, indicating that embedding drift alone does not determine the magnitude of decision impact.

\paragraph{Family-dependent coupling regimes.}
Different corruption families populate distinct regions of this space. JPEG compression (Fig.~\ref{fig:scatter_grid}b) exhibits a tight, approximately linear relationship between $\Delta$ and $D_{\Lambda}$, consistent with its high Spearman correlation in Table~\ref{tab:main-results}. In contrast, motion blur and occlusion (Fig.~\ref{fig:scatter_grid}c--d) display much broader vertical dispersion. For these families, similar levels of embedding drift can correspond to widely varying margin shifts, indicating weaker coupling between representation and decision.

\paragraph{Low-drift, high-impact regime.}
Of particular concern is the region characterized by relatively small $\Delta$ but large $D_{\Lambda}$. This regime appears prominently in motion blur and occlusion. Points in this region correspond to perturbations that minimally affect the global embedding while significantly altering the relative similarity to hazardous versus non-hazardous prompts. Empirically, these cases account for a large fraction of observed flips, suggesting that they represent a critical failure mode for VLM-based hazard detection.

\paragraph{Severity stratification.}
Coloring by severity reveals that embedding drift increases relatively smoothly with corruption strength, while margin drift exhibits more heterogeneous growth. In tightly coupled families (e.g., JPEG), increasing severity moves points outward along a narrow band. In weakly coupled families (e.g., occlusion), severity increases both the mean and the variance of margin drift, further amplifying unpredictability at higher corruption levels.

\paragraph{Interpretation.}
The scatter structure indicates that embedding drift constrains but does not determine decision change. While large margin shifts require non-negligible representation movement, similar levels of embedding drift can lead to substantially different margin outcomes depending on corruption type. This asymmetry highlights the role of corruption-specific effects in shaping decision stability.


\paragraph{Hypothesis: bounded decision sensitivity.}
The observed upper-envelope structure suggests a potential Lipschitz-like relationship between embedding drift and margin drift of the form $D_{\Lambda} \leq L \cdot \Delta$ for some task-dependent constant $L$. Our empirical geometry indicates that decision changes are indeed constrained by representation movement but not tightly determined by it.
Preliminary experiments using the Cauchy-Schwarz inequality (see Eq.~\ref{eq:upper_bound}) estimated the value of $L$ to be somewhere in the range $[0.70,1.10]$, depending on the choice of prompts, whereas visual inspection of Fig.~\ref{fig:scatter_grid} suggests that the boundary likely lies somewhere in the range of $[0.05,0.07]$. This suggests that the true slope $L$ of the tightest upper bound is an order of magnitude shallower than the Cauchy-Schwarz bound we had hypothesized.
Future work could aim to determine this tight upper-envelope constant analytically or empirically, potentially yielding formal guarantees on decision stability under bounded embedding perturbations.

\vspace{-0.22cm}

\subsection{Implications for Benchmarking}
These results suggest that robustness evaluation for multimodal systems should incorporate task-aligned metrics in addition to representation-level measures. In particular, reporting margin drift and threshold-flip rates by corruption family provides a more direct view of decision stability under shift. Benchmarks that rely solely on embedding perturbation statistics may overlook failure modes that have limited representation impact but significant decision consequences.

\vspace{-0.22cm}

\section{Conclusion}
We presented a compact, task-aligned stability analysis for CLIP-based hazard detection under controlled image corruption. Across seven corruption families, embedding drift and hazard-margin drift were related but far from equivalent. Some families produced predictable coupling between the two, while others created dangerous decision instability that was not well reflected by embedding movement alone. Additionally, corruption families differ not only in flip rate but in failure direction: most families are dominated by false negatives that suppress hazard detections, while occlusion produces predominantly false positives --- a divergence with direct implications for how corruptions should be weighted in safety-critical benchmarks.
For theory-informed benchmark design, this suggests that representation robustness should be complemented by task-aligned stability metrics that quantify how corruption propagates into decision variables and final verdicts.

\bibliography{references}
\bibliographystyle{IEEEtran}

\onecolumn

\appendix
\section{Prompt sets and corruption ranges}
\label{sec:prompt_set_appx}
Hazardous prompts: \emph{a pedestrian crossing the road}, \emph{a cyclist in the lane}, \emph{a stopped vehicle blocking traffic}, \emph{road debris on the street}, \emph{a construction zone on the road}, \emph{an emergency vehicle with flashing lights}. Non-hazardous prompts: \emph{a clear road with no obstacles}, \emph{a normal highway scene}, \emph{an empty intersection}, \emph{a clear lane ahead}, \emph{a normal driving scene}. Corruption severities are uniformly discretized over the same ranges used in the class draft: fog density $[0.15,0.75]$, Gaussian blur $\sigma\in[1,8]$, motion-blur kernel size $[3,25]$, JPEG quality $[90,10]$, low-light scale $[0.8,0.2]$, occlusion mask ratio $[0.05,0.60]$, and downsampling scale $[0.90,0.20]$.

\section{Lambda value visualization}
\label{sec:lambda_appx}
\begin{figure}
    \centering
    \includegraphics[width=1.00\linewidth]{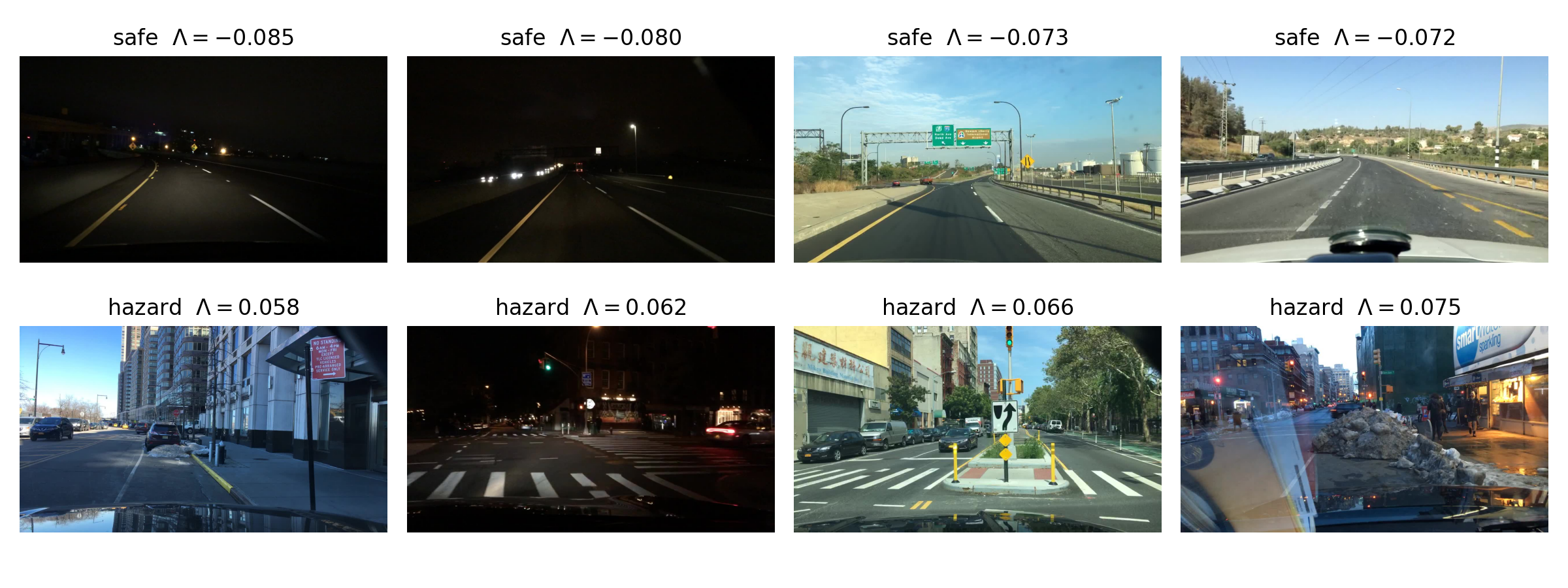}
    \caption{Top 4 most safe (top row) and hazardous (bottom row) images in the dataset, based on hazard score $\Lambda$.}
    \label{fig:lambda_scores}
\end{figure}

Fig.~\ref{fig:lambda_scores} shows the top 4 most and least hazardous scenes in the BDD100K dataset, according to our $\Lambda$ function using the prompt sets listed in Appendix~\ref{sec:prompt_set_appx}. Notice how the non-hazardous images are mostly clear roads with minimal traffic, while the hazardous images depict more busy, complicated scenarios.

\section{Corruption visualization}

\begin{figure}[h]
    \centering
    \includegraphics[width=1.02\linewidth]{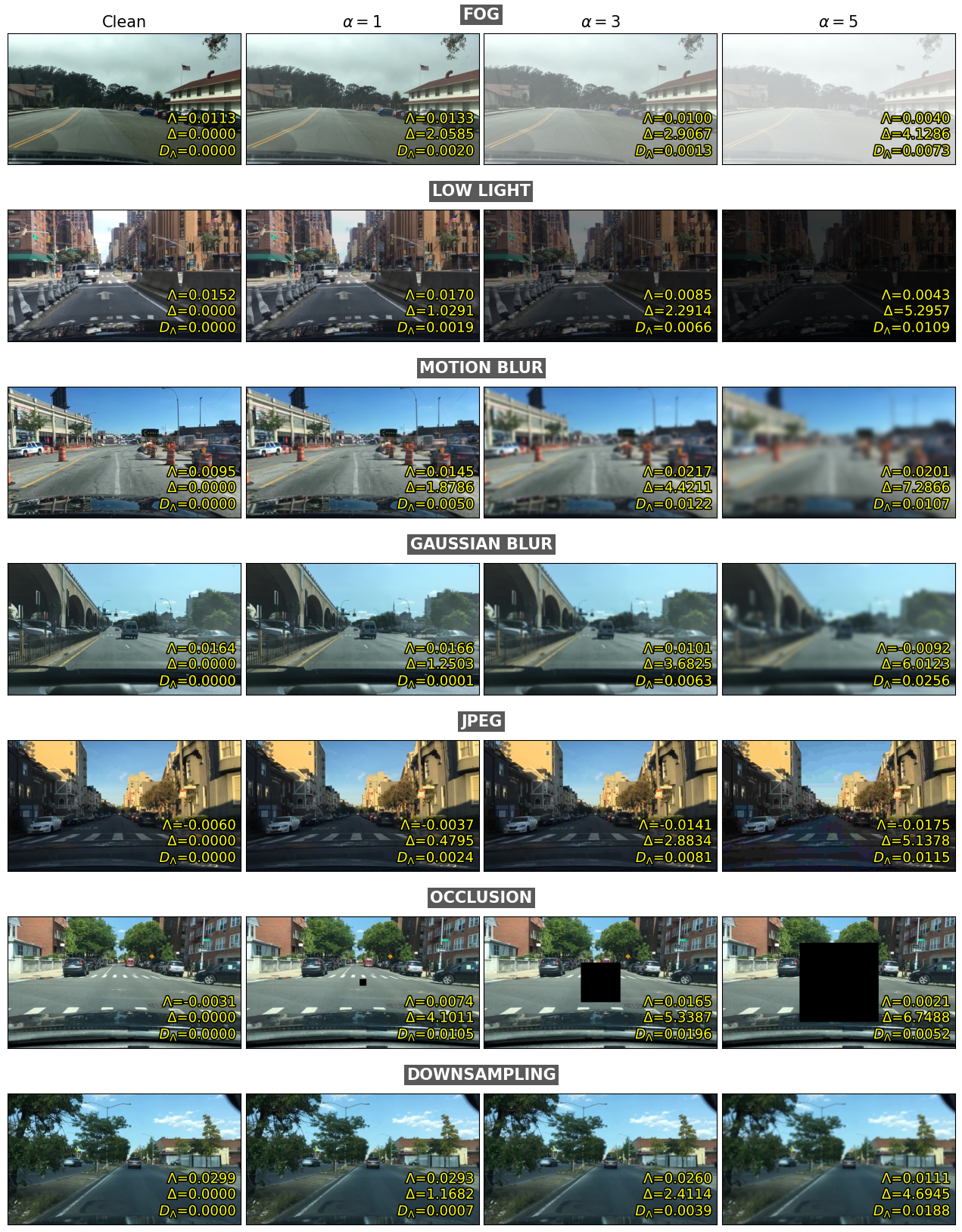}
    \caption{Visualizations of image corruptions on randomly-selected road images. Hazard level $\Lambda$, embedding drift $\Delta$, and margin drift $D_\Lambda$ are overlaid on each image.}
    \label{fig:corruption_vis}
\end{figure}

Fig.~\ref{fig:corruption_vis} illustrates each corruption operator being applied at various severity levels on real driving scenes. Each image is annotated with the corresponding hazard score ($\Lambda$), embedding drift ($\Delta$), and margin drift ($D_\Lambda$).

\end{document}